\documentclass{article}
\usepackage{stywhispers,amsmath,epsfig,subcaption,comment}
\usepackage{float}
\usepackage{tikz}
\usepackage[absolute,overlay]{textpos}
\usepackage{enumitem}
\usepackage{blindtext}
\usepackage{relsize}
\usepackage{hyperref}
\usepackage{url}

\tikzset{fontscale/.style={font=\relsize{#1}}}

\title{CHALLENGES IN HYPERSPECTRAL IMAGING FOR AUTONOMOUS DRIVING: THE HSI-DRIVE CASE}
\name{Koldo Basterretxea$^1$, Jon Gutiérrez-Zaballa$^2$, and Javier Echanobe$^2$ \thanks{This work was partially supported by the University of the Basque Country (UPV/EHU) under grant GIU21/007.}}
\address{1 University  of the Basque Country, Dep. of Electronics Technology\\
2 University  of the Basque Country, Dep. of Electricity and Electronics}
\begin{document}
%\ninept
%
\maketitle

\begin{textblock*}{21cm}(1.5cm,26cm)
\begin{tikzpicture}
    \draw (0,0) rectangle (17.5,0.5); 
    \end{tikzpicture}
\end{textblock*} 

\begin{textblock*}{21cm}(0cm,26cm)
\begin{tikzpicture}
    \node (center) {c};
    \path (center)+(10.5,4) node [fontscale=-1] (name) {\copyright 2025 IEEE. Final published version of the article can be found at \href{}{}.};
    \end{tikzpicture}
\end{textblock*} 

\begin{abstract}
The use of hyperspectral imaging (HSI) in autonomous driving (AD), while promising, faces many challenges related to the specifics and requirements of this application domain.
On the one hand, non-controlled and variable lighting conditions, the wide depth-of-field ranges, and dynamic scenes with fast-moving objects.
On the other hand, the requirements for real-time operation and the limited computational resources of embedded platforms.
The combination of these factors determines both the criteria for selecting appropriate HSI technologies and the development of custom vision algorithms that leverage the spectral and spatial information obtained from the sensors.
In this article, we analyse several techniques explored in the research of HSI-based vision systems with application to AD, using as an example results obtained from experiments using data from the most recent version of the HSI-Drive dataset.
\end{abstract}
\section{Introduction}
\label{sec:intro}

Recent advances in hyperspectral imaging (HSI) sensing technologies have enabled the commercialization of snapshot cameras capable of capturing spectral information across dozens or even hundreds of bands at video rates \cite{west2018commercial}.
These developments have sparked significant interest in exploring the potential of HSI in emerging application domains that demand affordable, compact, and portable spectral imaging systems with moderate to low power consumption such as precision agriculture, remote sensing in the new space, medical imaging for surgery, food processing, and autonomous navigation  \cite{Yako2025}.
In particular, some researchers have focused their attention on the potential of HSI to overcome some of the limitations of current intelligent vision systems for Autonomous Driving (AD), as well as to enhance their capability for comprehensive scene understanding \cite{gutierrez2023hsi}, \cite{10876494}.
However, the use of HSI snapshot cameras in AD requires a careful analysis of involved technologies and the customization of both algorithms and processing architectures to meet the specific requirements of this field of application.

The challenges associated with applying HSI to AD stem from the combination of factors that affect the quality and accuracy of collected data -constraints imposed by snapshot camera technologies, uncontrolled illumination, the presence of fast moving elements, the need for variable exposure times, etc.- together with the requirements of low-latency operation and the limitations in the computational power of available embedded processing platforms.
In this article, we share several lessons learned during the development of HSI-based image segmentation systems for AD using the HSI-Drive dataset.
We also describe recent improvements introduced in both the latest public release of this dataset and the deep neural network (DNN) image segmentation models developed through experiments with these data.

\section{HSI SNAPSHOT camera technologies}
\label{sec:camera}

The selection of an HSI camera to experiment with the development of advanced machine vision systems for AD must be grounded on a set of well established criteria.
Ideally,
\begin{itemize} 
    \item it must be capable of operating at video-rates (snapshot cameras) under different illumination conditions,
    \item it must be small, compact, and mechanically robust,
    \item it must provide enough spatial resolution for the depth of field required in the application,
    \item it must provide enough spectral resolution to achieve good spectral separability of the classes/materials to be identified,
    \item and it must be based on a scalable sensor technology that makes it competitive for future mass deployment.
\end{itemize}

Regarding the ability of HSI systems to operate at video rates, an additional factor that is often overlooked, but of considerable importance, is the processing required to reconstruct spatial and spectral information from the raw data.
A thorough analysis of commercially available HSI snapshot camera technologies is beyond the scope of this work.
However, it is worth noting that the main players in the small-form factor, high-throughput (\textgreater 10-20 fps) HSI snapshot camera segment employ either on-chip deposition technology of mosaics of narrow-band interferometric filters (e.g., Imec sensors integrated in Ximea and Photonfocus cameras), or light-field technology based on microlens arrays for multiple projection of spectrally filtered sub-images onto the sensor (as in the case of Cubert).
In both cases, the incident light is projected onto a standard CMOS sensor.
Alternative technologies, such as coded aperture snapshot spectral imaging (CASSI) combined with scattered spectral sampling, require complex and time-consuming data processing.
Consequently, these solutions are not yet capable of meeting the real-time performance requirements of autonomous driving applications.

The HSI-Drive dataset was generated from recordings with a Photonfocus MV2 camera featuring an Imec 25-band Red-NIR sensor with on-chip mosaic filter technology.
The selection of this system was motivated by several factors: its relatively low cost, its high throughput combined with acceptable spatial and spectral resolution, and the accessibility to the cube-generation processing code, which could be adapted and optimised for efficient execution on an embedded processing platform.
Moreover, spectral filter-on-chip technology offers far greater scalability than competing approaches, enabling mass production at costs comparable to those of traditional CMOS sensors.
Nonetheless, there are also some drawbacks to be considered when compared to other technological alternatives.
These include the presence of spectral leakage and second-order response peaks, the constraints imposed by mosaic patterns on the number of bands, and the need for pixel-level spatial realignment of the mosaic filters.
All in all, the design of the HSI-Drive dataset was  guided primarily by principles of simplicity and feasibility, rather than by the pursuit of maximum spectral quality.
In other words, our guiding question was: Given currently available and potentially scalable technologies for vehicle integration at reasonable cost, what level of useful information can be extracted from an HSI snapshot camera for the development of machine vision systems for autonomous driving?

\section{spectral information: data quality, consistency, and processing constraints}
\label{sec:spectral}

Considering that the approach for the development of the machine vision system under study, i.e. an image segmentation processing pipeline that could be executed on an embedded processing platform at video rates, was to be based on a machine learning (ML) model, to what extent the correspondence of the acquired spectral data with the real physical spectral reflectance signatures of the materials in a scene is of real importance? If that is the case, in principle, as far as there is data consistency, accuracy does not seem to be of any concern.
However, this is not entirely true, as it must be understood that the spectral separability between materials may depend on subtle differences in the spectral reflectance signatures.
However, here again the approach in the HSI-Drive dataset leans towards computational simplicity as far as the model quality is not compromised.

The latest version (v2.1) of the HSI-Drive dataset presents two major differences with respect to the previous v2.0 version: first, the annotation, while still favouring the preservation of the spectral information, has been carefully reviewed and the total number of labelled pixels has been increased with more than 1,100,000 new pixels (+2,5 \%).
Secondly, a new function to enhance data consistency has been added to the cube processing pipeline, which performs a pseudo-reflectance correction algorithm using data contained in the recorded frame itself.

\begin{figure}[H]
     \centering
     \begin{subfigure}[b]{\linewidth}
         \centering
         \includegraphics[width=0.74\linewidth]{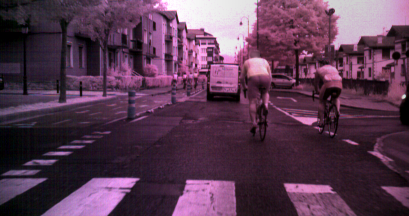}
         \caption{False RGB}
         \label{fig:RGB}
     \end{subfigure}
     \hfill
     \centering
     \begin{subfigure}[b]{\linewidth}
         \centering
         \includegraphics[width=0.95\linewidth]{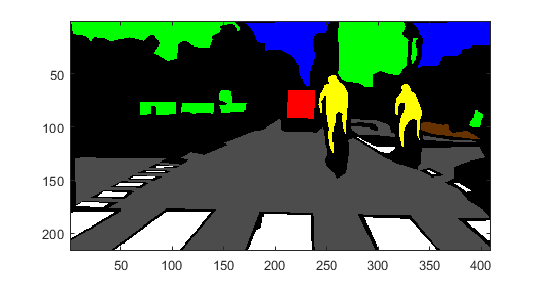}
         \caption{HSI-Drive 2.0 labelling}
         \label{fig:GT20}
     \end{subfigure}

     \vspace{0.2cm}
     \begin{subfigure}[b]{\linewidth}
         \centering
         \includegraphics[width=0.95\linewidth]{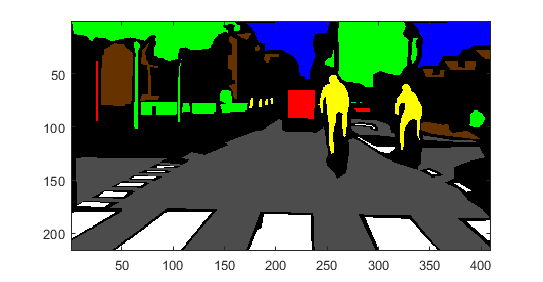}
         \caption{HSI-Drive 2.1 labelling}
         \label{fig:GT21}
     \end{subfigure}
        \caption{Example of manually labelled ground true images in HSI-Drive versions v2.0 and v2.1}
\end{figure}

\subsection{New data labeling}
\label{ssec:subhead}
The v2.1 version of the HSI-Drive dataset does not provide more annotated images, but a new, more careful annotation of the images already in version v2.0.
The aim of this new labelling effort has been twofold.
Firstly, to increase the amount of labelled pixels for training, especially in the most underrepresented categories.
Secondly, to provide higher quality test images for the evaluation of segmentation models.
However, the primary approach to the image labelling of the dataset is not changed, i.e. keeping unlabelled the pixels that a human labeller cannot clearly decide which category they belong to
This usually includes many background pixels and all the edges that delimit different items or surfaces in a scene.

\begin{table*}[h]
\centering
\caption{Frequency of each class in the HSI Drive v2.0 dataset.}
\label{tab:dataset2.0Partition}
\resizebox{18cm}{!}{%
\begin{tabular}{cccccccccccc}
  \cline{2-12}
  \multicolumn{1}{c|}{} &
  \multicolumn{1}{c|}{\textbf{Total}} &
  \multicolumn{1}{c|}{\textbf{Road}} &
  \multicolumn{1}{c|}{\textbf{R.Marks $^{\mathrm{a}}$}} &
  \multicolumn{1}{c|}{\textbf{Veg.$^{\mathrm{b}}$}} &
  \multicolumn{1}{c|}{\textbf{Pain.Met. $^{\mathrm{c}}$}} &
  \multicolumn{1}{c|}{\textbf{Sky}} &
  \multicolumn{1}{c|}{\textbf{Concrete}} &
  \multicolumn{1}{c|}{\textbf{Ped. $^{\mathrm{d}}$}} &
  \multicolumn{1}{c|}{\textbf{Water}} &
  \multicolumn{1}{c|}{\textbf{Unpain.Met. $^{\mathrm{e}}$}} &
  \multicolumn{1}{c|}{\textbf{Glass}} \\ \hline
  \multicolumn{1}{|c|}{\textbf{Pixels}} &
  \multicolumn{1}{c|}{43,947,503} &
  \multicolumn{1}{c|}{26,690,619} &
  \multicolumn{1}{c|}{1,325,343} &
  \multicolumn{1}{c|}{9,339,224} &
  \multicolumn{1}{c|}{948,852} &
  \multicolumn{1}{c|}{2,511,496} &
  \multicolumn{1}{c|}{2,315,153} &
  \multicolumn{1}{c|}{209,531} &
  \multicolumn{1}{c|}{12,330} &
  \multicolumn{1}{c|}{348,341} &
  \multicolumn{1}{c|}{246,614} \\ \hline
  \multicolumn{1}{|c|}{\textbf{\%}} &
  \multicolumn{1}{c|}{100} &
  \multicolumn{1}{c|}{60.73} &
  \multicolumn{1}{c|}{3.02} &
  \multicolumn{1}{c|}{21.25} &
  \multicolumn{1}{c|}{2.16} &
  \multicolumn{1}{c|}{5.71} &
  \multicolumn{1}{c|}{5.27} &
  \multicolumn{1}{c|}{0.48} &
  \multicolumn{1}{c|}{0.03} &
  \multicolumn{1}{c|}{0.79} &
  \multicolumn{1}{c|}{0.56} \\ \hline
  \multicolumn{12}{l}{$^{\mathrm{a}}$Road Marks. $^{\mathrm{b}}$ Vegetation. $^{\mathrm{c}}$ Painted Metal. $^{\mathrm{d}}$ Pedestrian. $^{\mathrm{e}}$ Unpainted Metal.}
\end{tabular}}
\end{table*}

\begin{table*}[h]
\centering
\caption{Frequency of each class in the HSI Drive v2.1 dataset.}
\label{tab:dataset2_1Partition}
\resizebox{18cm}{!}{%
\begin{tabular}{cccccccccccc}
  \cline{2-12}
  \multicolumn{1}{c|}{} &
  \multicolumn{1}{c|}{\textbf{Total}} &
  \multicolumn{1}{c|}{\textbf{Road}} &
  \multicolumn{1}{c|}{\textbf{R.Marks $^{\mathrm{a}}$}} &
  \multicolumn{1}{c|}{\textbf{Veg.$^{\mathrm{b}}$}} &
  \multicolumn{1}{c|}{\textbf{Pain.Met. $^{\mathrm{c}}$}} &
  \multicolumn{1}{c|}{\textbf{Sky}} &
  \multicolumn{1}{c|}{\textbf{Concrete}} &
  \multicolumn{1}{c|}{\textbf{Ped. $^{\mathrm{d}}$}} &
  \multicolumn{1}{c|}{\textbf{Water}} &
  \multicolumn{1}{c|}{\textbf{Unpain.Met. $^{\mathrm{e}}$}} &
  \multicolumn{1}{c|}{\textbf{Glass}} \\ \hline
  \multicolumn{1}{|c|}{\textbf{Pixels}} &
  \multicolumn{1}{c|}{45,055,512} &
  \multicolumn{1}{c|}{26,753,811} &
  \multicolumn{1}{c|}{1,364,908} &
  \multicolumn{1}{c|}{9,799,475} &
  \multicolumn{1}{c|}{1,113,573} &
  \multicolumn{1}{c|}{2,549,527} &
  \multicolumn{1}{c|}{2,485,658} &
  \multicolumn{1}{c|}{231,019} &
  \multicolumn{1}{c|}{10,592} &
  \multicolumn{1}{c|}{467,688} &
  \multicolumn{1}{c|}{279,261} \\ \hline
  \multicolumn{1}{|c|}{\textbf{\%}} &
  \multicolumn{1}{c|}{100} &
  \multicolumn{1}{c|}{59.38} &
  \multicolumn{1}{c|}{3.03} &
  \multicolumn{1}{c|}{21.75} &
  \multicolumn{1}{c|}{2.47} &
  \multicolumn{1}{c|}{5.66} &
  \multicolumn{1}{c|}{5.52} &
  \multicolumn{1}{c|}{0.51} &
  \multicolumn{1}{c|}{0.02} &
  \multicolumn{1}{c|}{1.04} &
  \multicolumn{1}{c|}{0.62} \\ \hline
  \multicolumn{12}{l}{$^{\mathrm{a}}$Road Marks. $^{\mathrm{b}}$ Vegetation. $^{\mathrm{c}}$ Painted Metal. $^{\mathrm{d}}$ Pedestrian. $^{\mathrm{e}}$ Unpainted Metal.}
\end{tabular}}
\end{table*}

\subsection{Reflectance correction and data normalization}
\label{ssec:reflectance}
With the aim of preserving the feasibility of real-time deployments, the cube processing pipeline applied to generate the hyperspectral cubes in the HSI-Drive 2.1 was kept simple as in previous versions.
The target application (AD) requires real-time processing of the acquired images and, considering the recording circumstances -outdoor recording, different camera setups, no additional lighting and spectra measurements etc., trying more accurate yet more complex cube generation pipelines would become the principal processing bottleneck with no noticeable improvements in the final results (accuracy).

The applied spectral cube generation process is a reflectance processing pipeline that comprehends the following steps
\begin{enumerate}[itemsep=2pt]
    \item Image cropping and framing.
    \item Bias removal and reflectance correction.
    \item Partial demosaicing (with 1/5 spatial resolution loss).
    \item Spatial filtering (optional).
    \item Translation to centre (band alignment by bilinear interpolation).
\end{enumerate}

Since data normalization techniques (per band normalization, pixel normalization etc.) have different objectives depending on the algorithms to be used to subsequently process the hyperspectral images, and since it is the final step of the processing pipeline, unlike in the v1.x versions of the dataset, we do not provide normalized cubes in v2.x versions.
This process, if necessary, is left to the dataset users.

\begin{comment}
    A dark field image and a white field image are used to compensate for sensor noise and perform white balancing.
    Ideally, this process should transform the irradiance information into a more or less accurate spectral reflectance information providing data consistency between images.
    However, this cannot be achieved with the recording conditions used to record the images in this dataset.
    The v2.1 version incorporates a new processing function to estimate the scene relative illumination with respect to the reference white field images and perform a scaling of the white balancing process accordingly.
\end{comment}

 \begin{figure}[H]
\centering
\includegraphics[width=0.45
\textwidth]{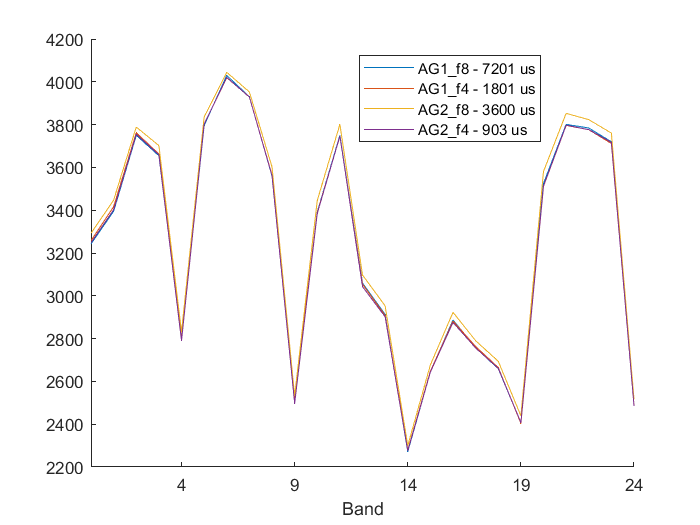}
\caption{figure}{Spectra of the maximum values of the averaged reference white tile images for the four different camera configurations used in the dataset.
Images were taken by exposing the calibrated tile to direct sunlight in a clear day with the sun at its zenith}
\label{fig:whites}
\end{figure}

Reflectance correction is aimed at cancelling the irradiance spectrum of the illuminant.
In a laboratory, this can be performed by using a calibrated reference white tile to divide the data in the image to be processed by the data in a reference white image acquired under the same illumination conditions and with the same camera configuration.
In dynamic outdoor conditions, as is the case for the recording of the HSI-Drive dataset scenes, an accurate reflectance correction is not possible at all.
However, when generating data for the previous versions of the HSI-Drive dataset, we chose to keep a "pseudo reflectance correction" stage by using a common white reference image obtained for each of the four different f-number/AG configurations of the camera.
These white reference images were generated by averaging various shots over a spectrally calibrated Spectralon white reflectance tile under expected natural maximum illumination conditions, i.e. at midday on a sunny day (Fig. \ref{fig:whites}).
Since no point spectrometer or other light measuring devices were present in the recording setup to compensate for illumination variations, obviously the applied correction does not fully normalize the spectral signatures of the different images in the dataset, and thus the data are not entirely consistent.
However, this processing still provides some benefits.
Firstly, it reduces sensor non-uniformity issues and image vignetting, and secondly, it cancels to some extent the irradiance spectrum of natural light.

Applying a posteriori data normalization techniques can reduce this issue to some extent.
The data used to train the CNN models for the HSI-Drive 2.0 experiments (see results published in \cite{gutierrez2023hsi}) were obtained after applying a per-pixel normalization of the spectral signatures.
This technique removes the irradiance offset produced by variations in incident light intensity.
The beneficial consequences are that shadow effects are mitigated -since reflectances on the same material surfaces are equalized- and that spectral information is favoured over the general reflectance of surfaces.
The negative effects are that the differences in the overall reflectance levels of different materials are removed, which results in the loss of valuable information for training AI models.

\begin{figure}[h]
     \centering
     \begin{subfigure}[b]{\linewidth}
         \centering
         \includegraphics[width=\linewidth]{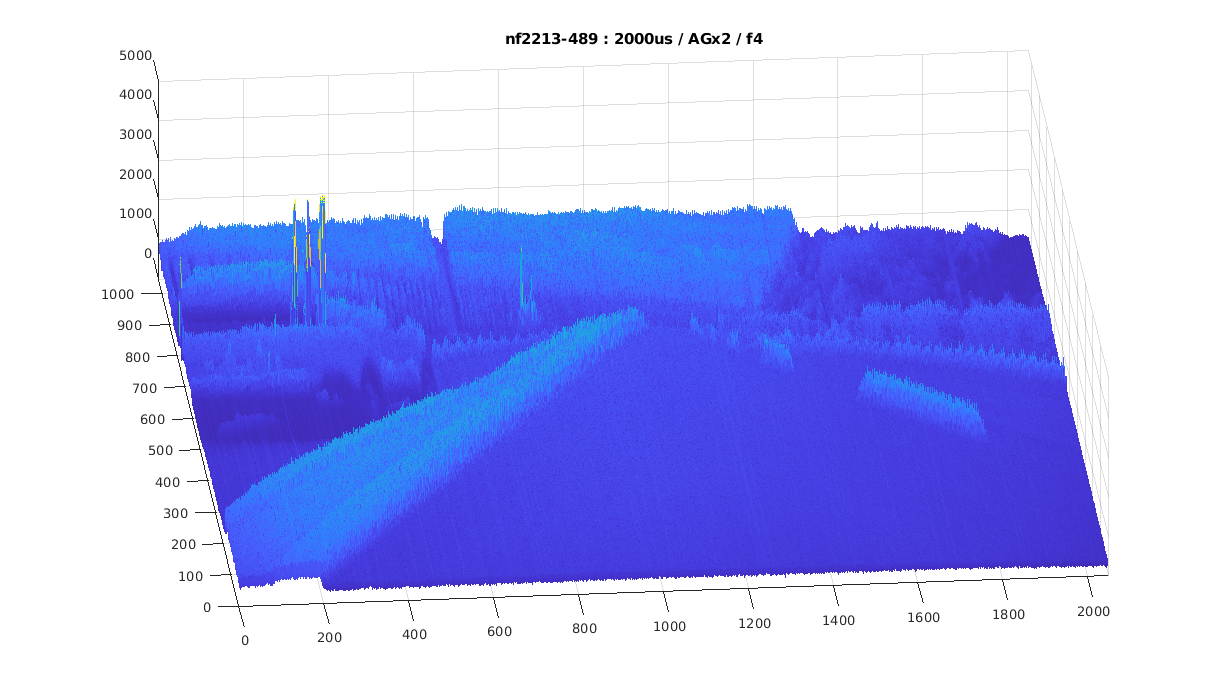}
         \caption{Sensed irradiance values (int12)}
         \label{fig:style 1 image a}
     \end{subfigure}
     \hfill
     \vspace{0.2cm}
     \begin{subfigure}[b]{\linewidth}
         \centering
         \includegraphics[width=0.75\linewidth]{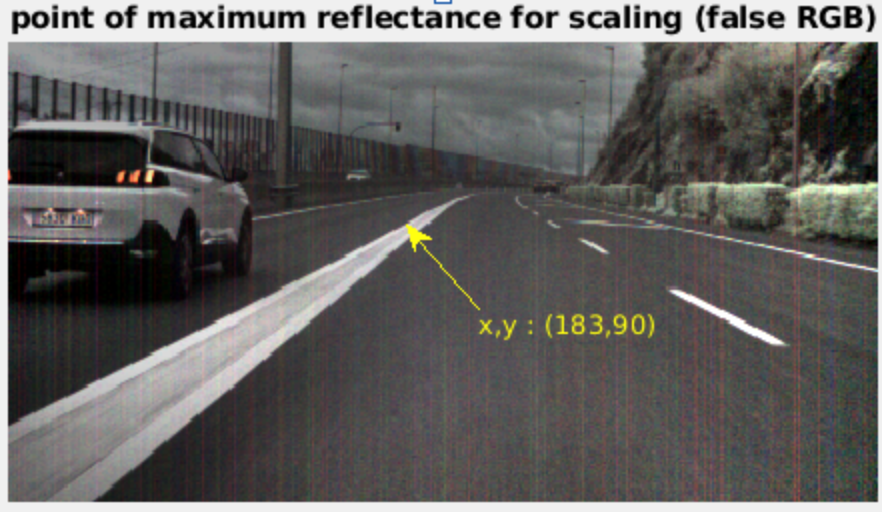}
         \caption{False RGB image of the 25-band HSI cube and coordinates of the pixel with maximum albedo}
         \label{fig:style 1 image b}
     \end{subfigure}
        \caption{Example of the identification of a maximum albedo pixel for the white balance scaling.
        This image corresponds to a cloudy Autumn morning recording with low lightning.
        Although the maximum irradiance values are generated by the rear and front lights of the cars (a), the algorithm successfully rejects those pixels and selects a pixel corresponding to the road mark as the highest reflectance pixel in the image (b)}
        \label{fig:bicycle}
\end{figure}

As an improvement to data quality, in the v2.1 version of the dataset we incorporate an additional processing function that estimates the relative level of illumination of the recorded scene by searching for the pixels with the highest albedo at each image.
These pixels usually correspond to high-reflectance white surfaces such as road marks, white vehicle bodies, etc., although in some cases the algorithm selects pixels corresponding to the sky.
By comparing the irradiance of these pixels with the reference white images, a scaling factor is calculated to correct the reference white images stored in memory.
Ideally, if the procedure was perfect, all images in the dataset would be scaled in the [0,1] range.
The search for reference pixels for scaling is not straightforward, since it involves rejecting pixels from artificial light sources such as illuminated signs, traffic lights, and front and rear lights of vehicles.
The programmed algorithm automatically segregates these "suspicious" pixels on the foundation of their spectral signatures, so no human intervention is required and thus, this function can be embedded in the image processing pipeline of the image segmentation processor (see Fig. \ref{fig:bicycle}).
Artificial light pixels are thus treated as outliers and clipped to 1 at the end of the cube preprocessing sequence.

\section{enhanced segmentation models and experimental results}
\label{sec:enhanced}

\subsection{U-Net with spectral attention modules}
\label{ssec:attention}

The U-Net trained with previous versions of HSI-Drive has been enhanced by incorporating attention modules.
These mechanisms are inspired by the human visual system, where the brain selectively prioritizes certain regions of the visual field while suppressing less relevant information.
In CNNs, a similar principle is implemented by weighting feature maps through an attention function.
This enables the network to emphasize discriminative spatial features, spectral features, or a combination of both, ultimately improving its representational capacity.

To leverage the spectral richness provided by hyperspectral images, several attention mechanisms have been investigated, namely Convolutional Block Attention Module (CBAM) \cite{woo2018cbamconvolutionalblockattention}, Squeeze-and-Excitation (SE) \cite{hu2018squeeze}, Efficient Channel Attention (ECA) \cite{wang2020ecanetefficientchannelattention}, and Coordinate Attention (CA) \cite{hou2021coordinateattentionefficientmobile}.
Among these, the best segmentation accuracy in our experiments was achieved with ECA which, additionally, introduces the least computational complexity and memory overhead to the model.

Efficient Channel Attention (ECA) operates by adaptively re-weighting channel-wise features without relying on dimensionality reduction.
Unlike SE, which compresses and then expands channel dimensions, ECA applies a local cross-channel interaction through a fast 1D convolution with a kernel size adaptively determined by the channel dimension.
This design ensures efficient information exchange between channels while avoiding additional fully connected layers, thus maintaining both accuracy and efficiency.

To effectively integrate ECA into the previously developed U-Net architecture, two attention blocks were incorporated at each encoder and decoder stage: one before the first convolutional block and another one before the second convolutional block (a detailed diagram of the original U-net model can be consulted in \cite{faults24}).
This placement allows the network to refine its feature representations at multiple depths, ensuring that both low-level and high-level spectral-spatial information are adaptively emphasized during the segmentation process.

\subsection{Experimental results}
\label{ssec:experimental}
Performed testing experiments on the HSI-Drive 2.0 and 2.1 datasets demonstrates the superiority of both the new modified U-Net model with spectral attention modules, and the new scaled reflectance correction processing over the use of non-scaled cubes both with and without pixel normalization.
Table \ref{table:inference_figures51} summarizes the results obtained in the HSI-Drive 2.0 dataset for a 5-class experiment with the previous U-Net using the pixel normalization technique.
All figures correspond to mean IoU values over a 5-fold cross-validation experimental setup.
Attention modules show over 2\% accuracy improvement in the weighted IoU index.
In Table \ref{table:inference_figures52} figures show a comparative study of the use of the new reflectance correction scaling algorithm with respect to the pixel normalization technique on the v2.1 dataset using the attention U-Net as predictor.
Here again, the improvement exceeds 2\% in accuracy.
Tables \ref{table:inference_figures61} and \ref{table:inference_figures61} show the results obtained for six class experiments by combining the attention modules with the new scaling reflectance correction technique.
The additional sixth classes, painted metal (vehicle bodywork, road signals etc.) and pedestrians/cyclists respectively, are specially challenging due to high intraclass spectral variability and low spectral interclass separability, as well as to small number of training data (see \ref{tab:dataset2_1Partition}).
The results obtained on these two classes are specially noteworthy, with accuracy improvements of 10.22\% and 5.09\% respectively.

Examples of segmented videos using these models can be found at \url{https://ipaccess.ehu.eus/HSI-Drive/.}

\begin{table}[h]
\centering
\caption{Segmentation results (\%) for the 5 class experiment on the HSI-Drive 2.0 dataset.}
\label{table:inference_figures51}
\resizebox{8.5cm}{!}{%
\begin{tabular}{|l|c|c|c|c|c|c|c|c|}
\hline
Model           &  Version          & road       & road m. & veg. & sky     & "others" & global  & weighted\\ \hline
U-Net           & No scaling+PN     & 97.53      & 85.94      & 95.04      &  93.02  & 78.59    &  94.64  & 87.52   \\ \hline
Att.U-Net       & No scaling+PN     & 98.05      & 87.74      & 95.54      &  95.25  & 82.94    &  95.64  & 89.71   \\ \hline
\end{tabular}}
\end{table}

\begin{table}[h]
\centering
\caption{Segmentation results (\%) for the 5 class experiment on the HSI-Drive 2.1 dataset.}
\label{table:inference_figures52}
\resizebox{8.5cm}{!}{%
\begin{tabular}{|l|c|c|c|c|c|c|c|c|}
\hline
Model           &  Version         & road & road m. & veg. & sky   & "others" & global & weighted\\ \hline
Att.U-Net       &  No scaling+PN   & 97.64& 85.33      & 94.55      & 92.89 & 81.79   &  94.71  & 87.75   \\ \hline
Att.U-Net       &  Scaling+PN      & 97.83& 87.27      & 94.60      & 94.14 & 82.50   &  95.04  & 89.16   \\ \hline
Att.U-Net       &  Scaling         & 98.04& 89.97      & 94.46      & 92.05 & 83.26   &  95.17  & 90.03   \\ \hline
\end{tabular}}
\end{table}

\begin{table}[H]
\centering
\caption{Segmentation results (\%) for the 6 class experiment (painted metal)}
\label{table:inference_figures61}
\resizebox{8.5cm}{!}{%
\begin{tabular}{|l|c|c|c|c|c|c|c|c|c|}
\hline
Model           &  Version         & road  & road m. & veg. &  p.metal &   sky   & "others" & global  & weighted \\ \hline
U-Net v2.0      &  No scaling+PN   & 97.34 & 85.20      & 93.84      &   58.61  &  92.30  & 68.65    &  93.07  & 74.45    \\ \hline
Att.U-Net v2.1  &  Scaling         & 98.08 & 90.34      & 93.63      &   68.83  &  91.44  & 74.61    &  93.97  & 81.09    \\ \hline
\end{tabular}}
\end{table}

\begin{table}[H]
\centering
\caption{Segmentation results (\%) for the 6 class experiment (pedestrians)}
\label{table:inference_figures62}
\resizebox{8.5cm}{!}{%
\begin{tabular}{|l|c|c|c|c|c|c|c|c|c|}
\hline
Model           &  Version         & road  & road m. & veg. &  ped. &   sky   & "others" & global  & weighted \\ \hline
U-Net v2.0      &  No scaling+PN   & 97.04 & 81.85      & 93.56      &   61.94      &  89.23  & 74.20    &  93.26  & 67.13    \\ \hline
Att.U-Net v2.1  &  Scaling         & 97.60 & 87.93      & 93.64      &   67.03      &  89.34  & 80.28    &  94.15  & 72.23    \\ \hline
\end{tabular}}
\end{table}

\section{concluding remarks}
\label{sec:conclusion}
The successful adoption of HSI technology in autonomous driving (AD) will depend on several key factors.
Firstly, it will depend on advances in HSI sensor technologies that enable the production of affordable yet technically precise snapshot cameras, capable of combining high image throughput with sufficient spectral and spatial resolution to support the development of high-performance machine vision systems for autonomous driving.
Secondly, it will depend on research into more capable and robust, yet computationally efficient algorithms that can make the most of the information provided by HSI data.
Finally, the combination of improvements achieved in both areas should lead to machine vision systems that demonstrate either their superiority or, at least, their complementarity to increasingly capable and precise systems based on more mature technologies.

In this paper, we share some research results obtained using the latest published version of the HSI-Drive dataset.
HSI-Drive is a dataset developed by recording real driving scenes with a single snapshot hyperspectral camera featuring a 25-band Red-NIR on-chip filter mosaic sensor.
In this version, we refined the labelling of ground-truth images, improved data consistency by introducing a customized illuminant intensity estimation algorithm for reflectance correction, and developed enhanced image segmentation models by incorporating spectral attention modules.
These additional lightweight attention blocks have been placed at key points in the encoder and decoder branches of the previous backbone U-net architecture to ensure efficient spectral information exchange between channels during inference.
The result is a consistent improvement in segmentation accuracy and robustness, while preserving processing simplicity for deployment on embedded devices.

\bibliographystyle{IEEEbib}
\bibliography{strings}

@Article{Yako2025,
author={Yako, Motoki},
title={Hyperspectral imaging: history and prospects},
journal={Optical Review},
year={2025},
month={Sep},
day={12},
issn={1349-9432},
doi={10.1007/s10043-025-01001-x},
url={https://doi.org/10.1007/s10043-025-01001-x}
}

@article{faults24,
title = {Evaluating single event upsets in deep neural networks for semantic segmentation: An embedded system perspective},
journal = {Journal of Systems Architecture},
volume = {154},
pages = {103242},
year = {2024},
issn = {1383-7621},
doi = {https://doi.org/10.1016/j.sysarc.2024.103242},
url = {https://www.sciencedirect.com/science/article/pii/S1383762124001796},
author = {Jon Gutiérrez-Zaballa and Koldo Basterretxea and Javier Echanobe},
}

@article{west2018commercial,
  title={Commercial snapshot spectral imaging: the art of the possible},
  author={West, Michael and Grossman, John and Galvan, Chris},
  year={2018}
}

@inproceedings{gutierrez2023hsi,
  title={{HSI-Drive v2.0: More data for new challenges in scene understanding for autonomous driving}},
  author={Guti{\'e}rrez-Zaballa, Jon and Basterretxea, Koldo and Echanobe, Javier and Mart{\'\i}nez, M Victoria and Martinez-Corral, Unai},
  booktitle={2023 IEEE Symposium Series on Computational Intelligence (SSCI)},
  pages={207--214},
  year={2023},
  organization={IEEE}
}

@INPROCEEDINGS{10876494,
  author={Shah, Imad Ali and Li, Jiarong and Glavin, Martin and Jones, Edward and Ward, Enda and Deegan, Brian},
  booktitle={2024 14th Workshop on Hyperspectral Imaging and Signal Processing: Evolution in Remote Sensing (WHISPERS)}, 
  title={Hyperspectral Imaging-Based Perception in Autonomous Driving Scenarios: Benchmarking Baseline Semantic Segmentation Models}, 
  year={2024},
  volume={},
  number={},
  pages={1-5},
  doi={10.1109/WHISPERS65427.2024.10876494}}

@inproceedings{hu2018squeeze,
  title={{Squeeze-and-Excitation Networks}},
  author={Hu, Jie and Shen, Li and Sun, Gang},
  booktitle={IEEE/CVF Conference on Computer Vision and Pattern Recognition},
  pages={7132--7141},
  year={2018},
  doi={10.1109/CVPR.2018.00745}
}

@INPROCEEDINGS{wang2020ecanetefficientchannelattention,
  author={Wang, Qilong and Wu, Banggu and Zhu, Pengfei and Li, Peihua and Zuo, Wangmeng and Hu, Qinghua},
  booktitle={IEEE/CVF Conference on Computer Vision and Pattern Recognition (CVPR)}, 
  title={{ECA-Net: Efficient Channel Attention for Deep Convolutional Neural Networks}}, 
  year={2020},
  pages={11531-11539},
  doi={10.1109/CVPR42600.2020.01155}}

@INPROCEEDINGS{hou2021coordinateattentionefficientmobile,  
  author={Hou, Qibin and Zhou, Daquan and Feng, Jiashi},
  booktitle={IEEE/CVF Conference on Computer Vision and Pattern Recognition (CVPR)}, 
  title={{Coordinate Attention for Efficient Mobile Network Design}}, 
  year={2021},
  pages={13708-13717},
  doi={10.1109/CVPR46437.2021.01350}}

@inproceedings{woo2018cbamconvolutionalblockattention,
  title={{CBAM: Convolutional Block Attention Module}},
  author={Woo, Sanghyun and Park, Jongchan and Lee, Joon-Young and Kweon, In So},
  booktitle={Computer Vision -- ECCV},
  publisher={Springer},
  pages={3--19},
  year={2018},
  doi={10.1007/978-3-030-01234-2_1}}

\end{document}